\documentclass[sigconf,nonacm]{acmart}

\setcopyright{acmlicensed}
\copyrightyear{2026}
\acmYear{2026}
\acmDOI{XXXXXXX.XXXXXXX}
\acmConference[MLCAD '26]{International Symposium on Machine Learning for CAD 2026}{September 7-9, 2026}{Jeju, South Korea}

\usepackage{amsmath}
\usepackage{graphicx}
\usepackage{svg}
\usepackage{float}
\usepackage{textcomp}
\usepackage{lipsum}
\usepackage{algorithm}     
\usepackage[mathscr]{euscript}
\usepackage{algpseudocode} 
\usepackage{xcolor}
\usepackage{multirow}
\usepackage{subcaption}
\usepackage{booktabs}
\usepackage{threeparttable}
\usepackage{tcolorbox}
\tcbuselibrary{skins}
\usepackage[dvipsnames]{xcolor}

\def\BibTeX{{\rm B\kern-.05em{\sc i\kern-.025em b}\kern-.08em
    T\kern-.1667em\lower.7ex\hbox{E}\kern-.125emX}}

\newboolean{anonymous}
\setboolean{anonymous}{false} 

\begin{document}

\title{Lighthouse RL: Sample-Efficient Circuit Optimization via Strategic Reset Points}

\ifthenelse{\boolean{anonymous}}{
  \author{Anonymous Authors}
  \affiliation{%
    \institution{Anonymous Institution(s)}
    \country{}}
  \renewcommand{\shortauthors}{Anonymous}
}{
    \author{Mustafa Emre Gürsoy\textsuperscript{1,2}, Stefan Uhlich\textsuperscript{1}, Ryoga Matsuo\textsuperscript{1,2}, Yağız Gençer\textsuperscript{1,2},\linebreak Arun Venkitaraman\textsuperscript{1}, Chia-Yu Hsieh\textsuperscript{1}, Andrea Bonetti\textsuperscript{1}, Eisaku Ohbuchi\textsuperscript{3}, Lorenzo Servadei\textsuperscript{1,4}}
    \affiliation{%
    \institution{\textsuperscript{1}\textit{Sony Group Corporation, Switzerland} \;
    \textsuperscript{2}\textit{EPFL, Switzerland} \;
    \textsuperscript{3}\textit{Sony Semiconductor Solutions, Japan} \;
    \textsuperscript{4}\textit{TU Munich, Germany}}\country{}}
    \renewcommand{\shortauthors}{Gürsoy, Uhlich, Matsuo, Gençer, Venkitaraman, Hsieh, Bonetti, Ohbuchi, Servadei}
}

\newcommand{\squeezeSpaceFigures}{\vspace{-0.1cm}}

\begin{abstract}
In this paper, we introduce Lighthouse RL, a sample-efficient reinforcement learning~(RL) approach for analog circuit sizing. Traditional methods lack generalization across different performance targets, while standard RL approaches waste resources exploring unpromising regions. Our method addresses these inefficiencies through a strategic reset strategy that initializes episodes from high-performing configurations discovered during training, called ``lighthouses''. These states, which are closer to the target objectives, guide exploration toward promising regions. When compared to RL and Bayesian optimization methods from the literature, we demonstrate the effectiveness of our approach on a 2D benchmark problem and on two analog circuits, showing significant improvements in sample efficiency~(up to 1.72× faster), optimization performance~(100\% vs. 0-87\% success rate), generalization~(75\% vs. 0-50\% extrapolation success), and objective maximization. This efficiency is particularly valuable for computationally expensive black-box optimization problems, and our reset strategy can be used as a plug-and-play enhancement for any RL-based optimization approach.
\end{abstract}


\maketitle

\section{Introduction}

Analog circuit sizing presents optimization challenges due to complex parameter-performance relationships. Traditional optimization approaches such as Bayesian Optimization~(BO)~\cite{shahriari2015taking_bo_simple} or Evolutionary Strategies~(ES)~\cite{beyer2002evolution_es1} require re-optimization from scratch whenever target objectives change. This limitation becomes increasingly problematic as the number of potential target combinations grows, making these methods computationally inefficient for practical circuit design workflows, where designers frequently need to explore multiple objectives. Although reinforcement learning~(RL) has shown promise for this task by learning adaptable policies, standard RL approaches still suffer from sample inefficiency, wasting computational resources exploring unpromising regions of the parameter space. In this paper, we introduce Lighthouse RL, a novel approach that addresses these limitations through a strategic reset strategy that adaptively initializes episodes from high-performing parameter configurations called ``lighthouses'' discovered during training. These states guide exploration toward promising regions by providing advantageous starting points closer to the target objectives. This approach significantly improves sample efficiency and generalization capabilities compared to conventional methods, making it particularly valuable for black-box circuit optimization problems where simulation could be computationally expensive.

\begin{figure}[!t]
    \centering
    \vspace{0.7cm}
    \includegraphics[width=\linewidth]{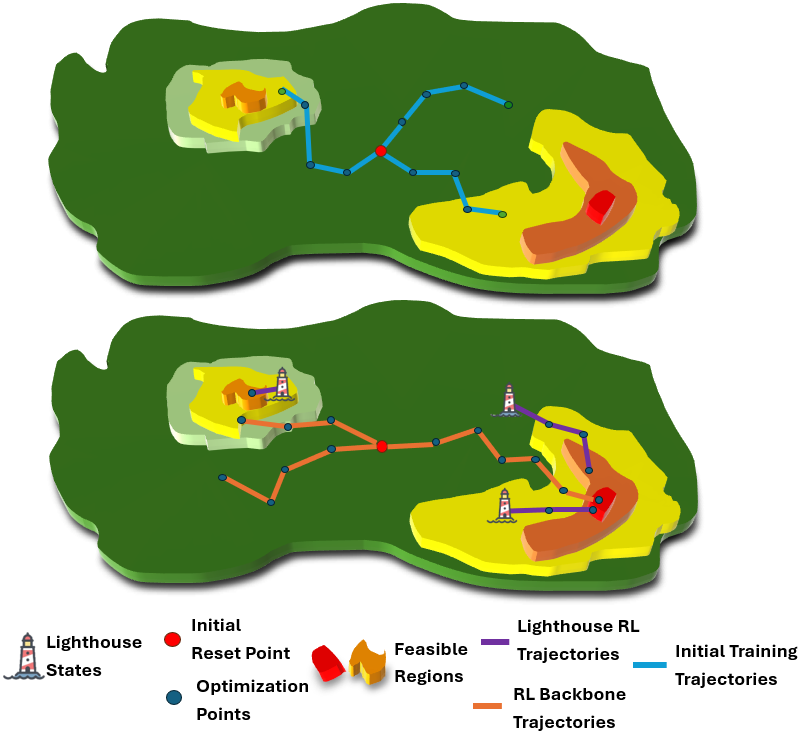}
    \caption{Concept of Lighthouse RL: strategic reset points guide exploration toward feasible solutions with fewer steps.}
    \label{fig:cartoon_illustration}
    \vspace{-0.3cm}
\end{figure}

In this work, we present a reinforcement learning formulation specifically designed for multi-objective circuit optimization problems with black-box evaluation functions. In particular, we present a \textit{novel reset strategy} for episodic reinforcement learning that maintains a set of high-performing solutions discovered during training. These states serve as strategic reset points for subsequent episodes, guiding exploration towards the promising regions of the state-space during training. The proposed reset strategy can be integrated with any RL-based black-box optimization approach without modifying the underlying algorithm, making our approach a \textit{plug-and-play enhancement} to existing algorithms.

\section{Background and Previous Work}
\subsection{Previous Work}
Circuit sizing optimization has evolved from traditional methods to machine learning approaches. Evolutionary algorithms~\cite{hakhamaneshi2019bagnet_es2, liu2009analog_es3} and Bayesian Optimization~\cite{lyu2018bo1, lyu2017bo2, touloupas2021bo3, liu2021parasiticbo4, touloupas2021locomobo_bo5} offer effective solutions but require re-optimization for each new target objective. Recent RL approaches by Settaluri et al.~\cite{settaluri2021automated_rl4} and Shi et al.~\cite{shi2022robustanalog_rl2} have shown promise in adaptability, while Wang et al.~\cite{wang2020gcn_rl5} combined graph neural networks with RL for transferable transistor sizing. Uhlmann et al.~\cite{uhlmann2022deep_rl6} addressed the challenge of sparse rewards in circuit sizing by implementing Hindsight Experience Replay~\cite{andrychowicz2017hindsight}. Budak et al.~\cite{budak2021dnn_rl3} explored changing reset points on a per-time step basis, which works well to maximize a scalar but struggles with generalization across objective ranges. Cao et al.~\cite{roseopt_rl1} improved sample efficiency by using Bayesian optimization to find starting points, but it is highly dependent on the quality of these initial solutions.

Despite these advances, existing approaches still struggle with sample efficiency when exploring the parameter space. Our work addresses this limitation through a strategic reset strategy that leverages high-performing configurations discovered during training, without requiring additional optimization methods.

\subsection{Reinforcement Learning}
Reinforcement learning~\cite{sutton1998reinforcement} enables agents to learn optimal decision making through environmental interaction, with successes in continuous control~\cite{tang2025deep_robotics}, games~\cite{mnih2013playingatarideepreinforcement_atari}, and autonomous driving~\cite{kiran2021deep_autonomous_driving}. The RL framework is typically expressed as a Markov Decision Process~(MDP) defined by $(\mathcal{S}, \mathcal{A}, \mathcal{P}, \mathcal{R}, \gamma)$, where $\mathcal{S}$ is the state space, $\mathcal{A}$ is the action space, $\mathcal{P}$ represents the transition probabilities, $\mathcal{R}$ is the reward function, and $\gamma \in [0,1]$ is the discount factor. The agent's goal is to learn a policy $\pi: \mathcal{S} \rightarrow \mathcal{A}$ that maximizes the expected cumulative discounted reward $J(\pi)$ by mapping states to actions. 

\subsection{Soft Actor-Critic}
Soft Actor-Critic~(SAC)~\cite{haarnoja2018soft} is an off-policy algorithm that optimizes both expected return and policy entropy for continuous control tasks. The algorithm employs three key neural networks: a policy network $\pi_\phi(a|s)$ parameterized by $\phi$, and two Q-networks $Q_{\theta_1}(s,a)$ and $Q_{\theta_2}(s,a)$ parameterized by $\theta_1$ and $\theta_2$.
The Q networks estimate the expected return and are trained to minimize the Bellman error with the loss function given as:
\begin{align}
    L_Q(\theta_i) &= \mathbb{E}_{(s,a,r,s') \sim \mathcal{D}} \Big[ \big( Q_{\theta_i}(s,a) - y \big)^2 \Big] \\
    y &= r + \gamma(Q_{\text{target}}(s',a') - \alpha \log \pi_\phi(a'|s'))
\end{align}
where $\quad a'\sim \pi_{\phi}(\cdot | s')$, $\mathcal{D}$ is the replay buffer, and $Q_{\text{target}}(s',a') = \min_{j=1,2} Q_{\bar{\theta}_j}(s',a')$ uses target networks to reduce the overestimation bias~\cite{haarnoja2018sactheoryandapplications}.

The policy network outputs a Gaussian distribution and is trained to maximize both Q-values and entropy with the objective of:
\begin{equation}
    \resizebox{0.90\linewidth}{!}{$
    J_{\text{SAC}}(\phi) = \mathbb{E}_{s \sim \mathcal{D}} \left[ \mathbb{E}_{a \sim \pi_\phi} \left[ \min_{i=1,2} Q_{\theta_i}(s,a) - \alpha \log \pi_\phi(a|s)  \right] \right]
    $}
\end{equation}
This dual objective allows SAC to balance exploitation~(maximizing Q-values) with exploration~(maximizing entropy), making it effective for complex continuous control problems. 

We chose SAC because, compared to on-policy methods like~\cite{sutton1999policyREINFORCE, schulman2017proximal, schulman2015trust}, it offers better sample efficiency by reusing past experiences and eliminates the need for complex trust region constraints. Against other off-policy algorithms such as~\cite{lillicrap2015continuousddpg, fujimoto2018addressingtd3}, SAC's entropy maximization provides more robust exploration, reduces sensitivity to hyperparameters, and helps prevent policy collapse to deterministic solutions, resulting in better stability and performance across diverse environments~\cite{haarnoja2018sactheoryandapplications}.

\section{Problem Formulation}

We formulate the multi-objective circuit sizing task as a feasibility problem. Let $\mathbf{x} \in \mathbb{R}^n$ represent the vector of decision variables constrained within the bounds $\mathbf{x}_{\min} \leq \mathbf{x} \leq \mathbf{x}_{\max}$. Let $\mathcal{I}$ denote the set of objective indices. For a given set of parameters $\mathbf{x}$, we define a set of objective functions $\mathbf{o}(\mathbf{x}) = [o_i(\mathbf{x})]_{i \in \mathcal{I}}$ that evaluate the quality of the solution. Each objective function $o_i(\mathbf{x})$ can be evaluated via a black-box process~(e.g., simulation), making gradient information unavailable. Our goal is to find a feasible solution $\mathbf{x}^*$, i.e., 
\begin{equation}
\text{Find } \mathbf{x}^* \in \mathcal{X} \quad \text{such that}  \quad o_i(\mathbf{x}^*) \geq o_{i,\text{target}} \quad \forall \, i \in \mathcal{I}
\end{equation}
where $\mathcal{X} = \{\mathbf{x} \in \mathbb{R}^n : \mathbf{x}_{\min} \leq \mathbf{x} \leq \mathbf{x}_{\max}\}$ represents the design space.

\textbf{Formulation of the RL problem:} We reformulate this feasibility problem as a reinforcement learning task. The RL agent operates in the continuous action space $\mathcal{A} \subset \mathbb{R}^n$ where each action $\mathbf{a} \in \mathcal{A}$ is bounded by predefined limits $[\mathbf{a}_{\min}, \mathbf{a}_{\max}]$.
At each time step $t$, the agent performs an incremental change to the parameters, followed by clamping to ensure the values remain within the designed space, that is:
\begin{equation}
    \mathbf{x}_{t+1} = \text{clip}(\mathbf{x}_t + \mathbf{a}_t, \mathbf{x}_{\min}, \mathbf{x}_{\max})
\end{equation}
where $\text{clip}(\mathbf{v}, \mathbf{v}_{\min}, \mathbf{v}_{\max})$ constrains each element of vector $\mathbf{v}$ to lie within the corresponding bounds. An episode consists of 30 time steps and is terminated early if the target objectives are met. 

\textbf{Reward function:} The total reward at each time step \( t \) encourages the agent to meet the target objectives.
The \textbf{objective penalty} \( r_{\text{obj}}(\mathbf{x}) \) penalizes undershooting targets:
\begin{equation}
    r_{\text{obj}}(\mathbf{x}) = \sum_{i \in \mathcal{I}} \alpha_i \cdot r_i(\mathbf{x})
\end{equation}
where for each objective, the reward component is defined as 
\begin{equation}
    r_i(\mathbf{x}) = \min(0,\frac{o_i(\mathbf{x}) - o_{i,\text{target}}}{o_i^{norm}})
\end{equation}
with \(\alpha_i\) being a weighting factor, and \( o_i^{\text{norm}} \) as a normalization constant.

The final reward at time step \( t \) is
\begin{equation}
    r_t(\mathbf{x_t}) = 
    \begin{cases}
        R &  \text{if } o_i(\mathbf{x}) \geq o_{i,\text{target}} \: \forall i \\
        \max(r_{\text{obj}(\mathbf{x_t})}, r_{\min}) & \text{otherwise}
    \end{cases}
\end{equation}
where $R$ is a substantial reward for the agent for finding a feasible solution, and $r_{\min}$ is the lower bound for stable learning. 

\textbf{Network structure:}
The network employs an actor-critic architecture that processes circuit structures as graphs, where each component~(e.g., NMOS, PMOS, resistor, capacitor) is represented as a node with features encoding component type and normalized parameters similar to~\cite{wang2020gcn_rl5}. The actor transforms the circuit graph through four Graph Attention Network~(GAT)~\cite{velivckovic2017graph_gat} layers with residual connections and layer normalization~\cite{ba2016layernorm}, while processing the design objectives separately via an MLP. These representations are concatenated and fed through another MLP to generate action distribution parameters. Critic functions similarly, but to obtain the Q-value, we concatenate the action with the objective and parameter embeddings after processing it via a different MLP.

\begin{algorithm}
\caption{Lighthouse RL}
\label{alg:sac_reset}
\scriptsize
\begin{algorithmic}
\State Initialize replay buffer $\mathcal{D}$, priority queue $\mathscr{P}$, successful set $\mathscr{S}$, lighthouse $\mathcal{L}$, parameters $n, m, k, N_{\text{update}}$

\begin{tcolorbox}[colback=teal!10,boxrule=0pt,left=0pt,right=0pt,top=0pt,bottom=0pt]
\State{\textcolor{gray}{\emph{// Phase 1: Exploration - Finding Lighthouse States}}}
\While{True}
    \If{episode mod $N_{\text{update}} = 1$}
        \State Sample new target $\mathbf{o}_{\text{target}}^{\text{new}} \sim \mathcal{U}(\mathbf{o}^{\min}, \mathbf{o}^{\max})$
        \State Form $\mathcal{R}$ as top $n$ params closest to $\mathbf{o}_{\text{target}}^{\text{new}}$ from $\mathscr{P}$
    \EndIf

    \State Reset env with randomly sampled $\mathbf{x}_0$ from $\mathcal{R}$, set target $\mathbf{o}_{\text{target}}^{\text{new}}$

    \For{each time step}
        \State Sample action $\mathbf{a}_t \sim \pi_\theta(\cdot | \mathbf{s}_t)$, execute, store transition, update model following SAC
    \EndFor

    \State Extract best parameters $\mathbf{x}^*$, specs $\mathbf{o}^*$

    \If{$\mathbf{o}^* \geq \mathbf{o}^{\min}$} 
        \State Add $(\mathbf{x}^*, \mathbf{o}^*)$ to $\mathscr{S}$
    \Else 
        \State Add $(d(\mathbf{o}^*, \mathbf{o}^{\max}), \mathbf{x}^*, \mathbf{o}^*)$ to $\mathscr{P}$
    \EndIf

    \If{$|\mathscr{S}| \geq m$}
        \State Choose top $k$ from $\mathscr{P}$ as lighthouse $\mathcal{L}$
        \State Break
    \EndIf
\EndWhile
\end{tcolorbox}
\squeezeSpaceFigures
\end{algorithmic}

\begin{algorithmic}[0]
\begin{tcolorbox}[colback=violet!10,boxrule=0pt,left=0pt,right=0pt,top=0pt,bottom=0pt]
\State{\textcolor{gray}{\emph{// Phase 2: Exploitation - Using Lighthouse States}}}
\While{True}
    \If{episode mod $N_{\text{update}} = 1$}
        \State Sample new target $\mathbf{o}_{\text{target}}^{\text{new}} \sim \mathcal{U}(\mathbf{o}^{\min}, \mathbf{o}^{\max})$
    \EndIf

    \State Reset env with randomly sampled $\mathbf{x}_0$ from $\mathcal{L}$, set target $\mathbf{o}_{\text{target}}^{\text{new}}$

    \For{each time step}
        \State Sample action $\mathbf{a}_t \sim \pi_\theta(\cdot | \mathbf{s}_t)$, execute, store transition, update model following SAC
    \EndFor

    \If{current time step $>$ maximum} \textbf{break} \EndIf
\EndWhile
\end{tcolorbox}
\end{algorithmic}
\vspace{-0.1cm}
\end{algorithm}

\section{Proposed method: Lighthouse RL}

Lighthouse RL addresses a fundamental challenge in reinforcement learning for optimization: balancing exploration of the parameter space with efficient exploitation of promising regions. As illustrated in Fig.~\ref{fig:cartoon_illustration}, traditional RL approaches typically start each episode from a fixed point~(red circle in top panel), requiring the agent to repeatedly explore similar trajectories~(blue/orange paths) to reach feasible solutions. However, learning in green regions is ineffective for optimization, adding only computational cost without benefit.

Our key insight is to strategically identify and leverage high-performance parameter configurations discovered during training as \textit{lighthouse states}~(bottom panel). These \textit{lighthouses} serve as intelligent starting points for subsequent episodes, allowing the agent to begin exploration from advantageous positions closer to feasible regions. 

The complete algorithm is presented in Alg.~\ref{alg:sac_reset}, which consists of two distinct phases: an exploration phase to discover these lighthouse states, followed by an exploitation phase that leverages them to accelerate convergence. We now describe each phase in detail.

\begin{figure*}[!t]
    \centering
    \begin{subfigure}[t]{0.25\textwidth}
        \centering
        \raisebox{0.45cm}{\includesvg[width=\linewidth]{figures/rewards_comparison_sphere.svg}}
        \caption{}
        \label{fig:benchmark_reward}
    \end{subfigure}
    \begin{subfigure}[t]{0.73\textwidth}
        \centering
        \includegraphics[width=\linewidth]{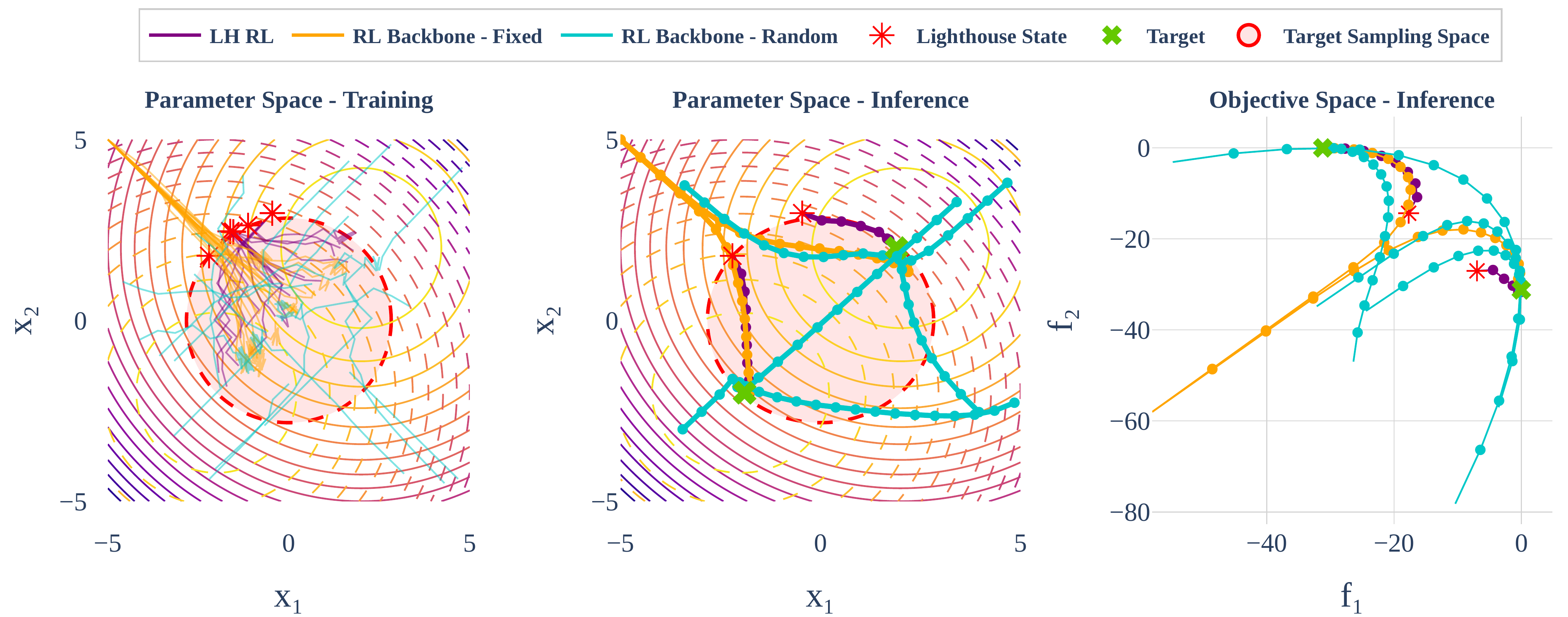}
        \caption{}
        \label{fig:bencmark_traj}
    \end{subfigure}
    \vspace{-0.3cm}
    \caption{(a) Total number of successful episodes over 12,000 training time steps. (b) Multi-objective optimization comparison between Lighthouse RL and RL backbone approaches with fixed and random reset points. (Left) Parameter space trajectories during training showing how each method navigates the search space. Lighthouse states provide strategic starting points that accelerate convergence to optimal solutions. (Middle) Parameter space trajectories during inference for two distinct targets, demonstrating Lighthouse RL's superior convergence properties. For the random reset RL backbone, several trajectories are presented starting from random reset points. (Right) Objective space visualization during inference showing how all three methods perform against target objectives. While all approaches successfully reach the inference target, Lighthouse RL takes the fewest steps.}
    \label{fig:combined_benchmark}
    \squeezeSpaceFigures
\end{figure*}

\subsection{Phase 1: Exploration - Finding Lighthouse States}

For each objective $i \in \mathcal{I}$, we define a target range $[o_i^{\text{min}}, o_i^{\text{max}}]$ representing acceptable performance bounds. During the exploration phase, the algorithm systematically searches for parameter configurations that approach but do not quite satisfy all minimum objectives. These near-optimal configurations, which we call \textit{lighthouse states}, will later serve as strategic starting points for the exploitation phase.
To use later, we define a distance function $d(\mathbf{o}_a, \mathbf{o}_b)$ between the objective vectors, 
\begin{equation}
d(\mathbf{o}_a, \mathbf{o}_b) = \sqrt{\frac{1}{|\mathcal{I}|} \sum_{i \in \mathcal{I}} \left(\frac{\min(0, o_{b,i} - o_{a,i})}{o_{i}^{\text{norm}}}\right)^2}
\end{equation}
where $o_{a,i}$ is the value of the objective $i$ in the vector $\mathbf{o}_a$, and $o_{i}^{\text{norm}}$ is a normalization factor for each objective.

At the beginning of each episode, we sample a new objective vector
$
\mathbf{o}_{\text{target}}^{\text{new}} \sim \mathcal{U}(\mathbf{o}^{\text{min}}, \mathbf{o}^{\text{max}})
\label{eq: target_sample}
$
to use as the next target.
To efficiently track our progress, we maintain two distinct sets of solutions.
First, the priority queue $\mathscr{P}$ is defined as 
\begin{align}
\mathscr{P} = \{(& d(\mathbf{o}_p, \mathbf{o}^{\text{max}}), \mathbf{x}_p, \mathbf{o}_p) \mid 
p \in \{1,2,...,|\mathscr{P}|\}, \nonumber\\
& \exists i \in \mathcal{I} : o_{p,i} < o_{i}^{\text{min}} \}
\end{align}
stores promising but incomplete solutions, where $d(\mathbf{o}_p, \mathbf{o}^{\text{max}})$ measures how close each solution is to the upper performance bounds, and sorts the queue. Solutions with smaller distance values are prioritized as they are closer to satisfying all objectives at their maximum levels.
Second, the successful set $\mathscr{S}$ is defined as
\begin{equation}
\mathscr{S} = \{(\mathbf{x}_s, \mathbf{o}_s) \mid s \in \{1,2, \dots |\mathscr{S}|\}, \forall i \in \mathcal{I} : o_{s,i} \geq o_{i}^{\text{min}}\}
\end{equation}
contains solutions that fully satisfy all minimum objectives.

This separation into $\mathscr{P}$ and $\mathscr{S}$ offers three key advantages:

\begin{enumerate}
    \item It prevents the waste of computational resources in initializing episodes with parameters that already satisfy all requirements.
    \item By tracking successful solutions in $\mathscr{S}$, we establish a clear criterion for when to transition from exploration to exploitation.
    \item The discovered solutions have practical value as ready-to-use parameter sets for analog designers with similar objectives.
\end{enumerate}

In every $N_{\text{update}}$ episodes, we sample new target objectives and select the $n$ parameters from $\mathscr{P}$ with minimal $d(\mathbf{o}_p, \mathbf{o}_{\text{target}}^{\text{new}})$ to form the reset point set $\mathcal{R}$. For each subsequent episode, we randomly sample a starting point from $\mathcal{R}$, focusing exploration on promising regions of the parameter space.

When the successful set $\mathscr{S}$ reaches a predefined size $m$, we select the first $k$ solutions of $\mathscr{P}$ with minimal $d(\mathbf{o}_s, \mathbf{o}^{\text{max}})$ as our lighthouse states $\mathcal{L} = \{\mathbf{x}_1^L, \mathbf{x}_2^L, ..., \mathbf{x}_k^L\}$. This selection marks the transition from the exploration phase to the exploitation phase, where we consolidate on a set of \textit{lighthouses}, and fine-tune the policy starting only from them.

\subsection{Phase 2: Exploitation - Using Lighthouse States}

In the exploitation phase, we stop the process of appending solutions to the sets $\mathscr{P}$ and $\mathscr{S}$. Instead, reset states are only randomly selected from the lighthouse states, i.e.:
\begin{equation}
\mathbf{x}_{\text{reset}} \sim \text{Uniform}(\mathcal{L})
\end{equation}
This randomization is important, as it gives the model the ability to generalize, rather than overfitting to a certain trajectory.

Phase 2 serves an important purpose in our approach. Once we have collected enough successful solutions in Phase 1, we will have sufficient lighthouse states to guide future exploitation. At this point, continuing to discover new regions would force the agent to adapt to foreign territories, which, while favorable for the diversity of solutions, would slow the training progress. By discontinuing the exploration of new parameter regions and focusing solely on the lighthouse states, we transition from exploration to exploitation, strengthening the model's understanding of the regions near these strategic starting points.

Although our formulation focuses on circuit optimization, the Lighthouse RL approach applies to general black-box optimization problems. In~\ref{sec:2d_benchmark}, we demonstrate this adaptability by studying a simple 2D multi-objective feasibility problem before addressing two more complex circuit sizing tasks.

\section{Experiments}

\subsection{2D Benchmark Function Analysis}
\label{sec:2d_benchmark}

To evaluate our reset strategy, we apply Lighthouse RL to a multi-objective feasibility problem featuring two sphere functions. Each sphere function is defined as $f_i(\mathbf{x}) = -\|\mathbf{x} - \mathbf{x}_i^0\|^2$ with global maxima at different locations: $\mathbf{x}_1^0 = -\mathbf{x}_2^0= [2, 2]$. This creates a landscape with two distinct regions of interest. The objective vector is defined as $[f_1(\mathbf{x}), f_2(\mathbf{x})]$, where we aim to maximize both function values.

Following our methodology, we sample the target objectives and assign the agent to find parameters $[x_1, x_2]$ that meet these objectives. After training our Lighthouse RL method and RL backbones with fixed and random reset points for 12,000 time steps, we evaluated them targeting $[1.5, 2]$ with function value $(-0.25, -28.25)$, a challenging point near the maxima of one of the sphere functions. As shown in Fig.~\ref{fig:bencmark_traj}, Lighthouse RL converges to target objectives in fewer inference steps than both alternatives. By initializing from lighthouse states rather than exploring suboptimal regions, Lighthouse RL reduces computational cost, which is particularly valuable when function evaluations are expensive.

As shown in Fig.~\ref{fig:benchmark_reward}, the graph indicates that Lighthouse RL achieves more successful episodes during training by leveraging previously discovered parameter sets. This benchmark demonstrates how Lighthouse RL navigates optimization landscapes with multiple regions of interest. This behavior is desirable, as many black-box optimization problems have multiple viable solutions that exist in different regions of the parameter space.

\begin{table}[!t]
\centering
\caption{Parameter Number in RL-Based Sizing Algorithms}
\label{tab:rl_circuit_params}
\renewcommand{\arraystretch}{0.6}
\setlength{\tabcolsep}{3pt}
\scriptsize
\begin{tabular}{@{}lccccccc@{}}
\toprule
\textbf{Reference} & \cite{roseopt_rl1} & \cite{shi2022robustanalog_rl2} & \cite{budak2021dnn_rl3} & \cite{settaluri2021automated_rl4} &\cite{wang2020gcn_rl5}& \textbf{This work} \\
\midrule
\textbf{\# Parameters} & 10 & 20 & 20 & \textbf{29} & 25 & \textbf{29} \\
\bottomrule
\end{tabular}
\vspace{-0.1cm}
\end{table}

\begin{table}[!t]
\centering
\caption{Ranges of Design Parameters and Objectives}
\label{tab:design_spec_ranges}
\renewcommand{\arraystretch}{0.9}
\setlength{\tabcolsep}{4pt}
\scriptsize
\begin{tabular}{@{}lcc|ccc@{}}
\toprule
\textbf{Parameters} & \multicolumn{2}{c|}{\textbf{Two-Stage}} & \multicolumn{3}{c}{\textbf{Multistage~\cite{qu2016design_big_opamp}}} \\
\multicolumn{6}{@{}l}{} \\
$W$ & \multicolumn{2}{c|}{1--64\,$\mu$m} & \multicolumn{3}{c}{0.36/0.42--100\,$\mu$m (NMOS/PMOS)} \\
$L$ & \multicolumn{2}{c|}{1--10\,$\mu$m} & \multicolumn{3}{c}{0.15--10\,$\mu$m} \\
$C$ & \multicolumn{2}{c|}{0.1--5\,nF} & \multicolumn{3}{c}{0.1--5\,nF} \\
$R$ & \multicolumn{2}{c|}{1--100k\,$\Omega$} & \multicolumn{3}{c}{0.1--1M\,$\Omega$} \\
VCM & \multicolumn{2}{c|}{--} & \multicolumn{3}{c}{0--1.8\,V} \\
I & \multicolumn{2}{c|}{--} & \multicolumn{3}{c}{1--100\,$\mu$A} \\
\# params & \multicolumn{2}{c|}{11} & \multicolumn{3}{c}{29} \\
\midrule
\multicolumn{6}{@{}l}{} \\
\textbf{Objectives} & \textbf{Train} & \textbf{Extrap.} & \textbf{Train} & \textbf{Extrap.} & \textbf{Max} \\
Gain & [60,65]\,dB & [65,70]\,dB & [55,65]\,dB & [75,80]\,dB & [130,135]\,dB \\
BW & [2,3]\,MHz & [3,4]\,MHz & [4,5]\,MHz & [5,6]\,MHz & [15,20]\,MHz \\
PM & [45,60]° & [60,70]° & [50,70]° & [65,70]° & [60,65]° \\
GM & [10,15]\,dB & [15,17.5]\,dB & [10,15]\,dB & [10,15]\,dB & [45,50]\,dB \\
Sat. V & $>$0V & $>$0V & - & - & - \\
\bottomrule
\end{tabular}
\vspace{-0.3cm}
\end{table}

\newcommand{\bestresult}[1]{\textbf{#1}}

\newcommand{\secondbest}[1]{#1}

\begin{table*}[!t]
\footnotesize
\begin{threeparttable}
\centering
\caption{Comparison of circuit sizing algorithms for two circuits.
Success rate and average inference steps are presented with means and one standard deviation for 5 different seeds.
The best results are in \bestresult{bold}.}
\label{tab:comprehensive_comparison}
\renewcommand{\arraystretch}{1.0}
\setlength{\tabcolsep}{3pt}
\begin{tabular*}{\textwidth}{@{\extracolsep{\fill}}l|ccc|cc|ccc|cc}
\toprule
& \multicolumn{5}{c|}{\textbf{Two-Stage OpAmp}} & \multicolumn{5}{c}{\textbf{Multistage Amplifier\cite{qu2016design_big_opamp}}} \\
\cmidrule(lr){2-6} \cmidrule(lr){7-11}
\multirow{2}{*}{\textbf{Method}} 
& \multicolumn{3}{c|}{\textbf{Within Distribution (C)}} 
& \multicolumn{2}{c|}{\textbf{Extrapolation (D)}}
& \multicolumn{3}{c|}{\textbf{Within Distribution (C)}} 
& \multicolumn{2}{c}{\textbf{Extrapolation (D)}} \\
\cmidrule(lr){2-4} \cmidrule(lr){5-6} \cmidrule(lr){7-9} \cmidrule(lr){10-11}
 & \textbf{T.S.E.}$^{\dagger}$ & \textbf{SR (\%)}$^{\ddagger}$ & \textbf{Inf. Steps}$^{\S}$ 
 & \textbf{SR (\%)} & \textbf{Steps}
 & \textbf{T.S.E.}$^{\dagger}$ & \textbf{SR (\%)}$^{\ddagger}$ & \textbf{Inf. Steps}$^{\S}$ 
 & \textbf{SR (\%)} & \textbf{Steps} \\
\midrule
\texttt{BO} & N/A & 37.6 $\pm$ 4.8 & 123.1 $\pm$ 6.2 & 8.0 $\pm$ 4.4 & 144.7 $\pm$ 2.8 & N/A & \secondbest{26.4 $\pm$ 6.2} & 560.8 $\pm$ 19.7 & \secondbest{10.0 $\pm$ 1.8} & 612.5 $\pm$ 2.0 \\
\texttt{RoSeOpt} & $1.00\times$ & 58.2 $\pm$ 15.4 & 18.0 $\pm$ 4.2 & 5.6 $\pm$ 4.0 & 28.7 $\pm$ 1.0 & \secondbest{$1.02\times$} & 13.2 $\pm$ 26.4 & \secondbest{28.0 $\pm$ 4.0} & 7.6 $\pm$ 15.2 & \secondbest{28.8 $\pm$ 2.3} \\
\texttt{RoSeOpt (f-g)} & \secondbest{$1.35\times$} & \secondbest{87.2 $\pm$ 14.7} & \secondbest{11.1 $\pm$ 1.2} & 10.0 $\pm$ 14.4 & 27.8 $\pm$ 3.1 & $1.00\times$ & 2.0 $\pm$ 4.0 & 29.8 $\pm$ 0.3 & 0.4 $\pm$ 0.3 & 29.9 $\pm$ 0.02 \\
\texttt{RL Backbone} & $1.10\times$ & \bestresult{100.0 $\pm$ 0.0} & 15.7 $\pm$ 2.1 & \secondbest{50.4 $\pm$ 41.6} & \secondbest{24.7 $\pm$ 5.6} & $1.00\times$ & 0.0 $\pm$ 0.0 & 30.0 $\pm$ 0.0 & 0.0 $\pm$ 0.0 & 30.0 $\pm$ 0.0 \\
\texttt{Lighthouse RL} & \bestresult{$\mathbf{1.55\times}$} & \bestresult{100.0 $\pm$ 0.0} & \bestresult{3.8 $\pm$ 1.4} & \bestresult{75.2 $\pm$ 26.8} & \bestresult{15.6 $\pm$ 6.5} & \bestresult{$\mathbf{1.72\times}$} & \bestresult{87.2 $\pm$ 18.0 } & \bestresult{7.9 $\pm$ 4.8} & \bestresult{29.2 $\pm$ 37.0} & \bestresult{22.8 $\pm$ 10.2} \\
\bottomrule
\end{tabular*}
\begin{tablenotes}\footnotesize
\item[$\dagger$] Training Sample Efficiency - relative training efficiency~(number of simulations to reach 50 successful episodes), normalized to the least efficient method.
\item[$\ddagger$] Success rate - percentage of successful episodes during inference out of 50 randomly sampled target objectives.
\item[$\S$] Inference Steps -  average number of simulations for inference episodes. For failed episodes, the maximum episode length was used.
\end{tablenotes}
\end{threeparttable}
\squeezeSpaceFigures
\end{table*}

\subsection{Analog Circuit Sizing: Experimental Setup}

\begin{figure}[!t]
    \centering
    \includegraphics[width=0.6\linewidth,trim=0 10 0 0]{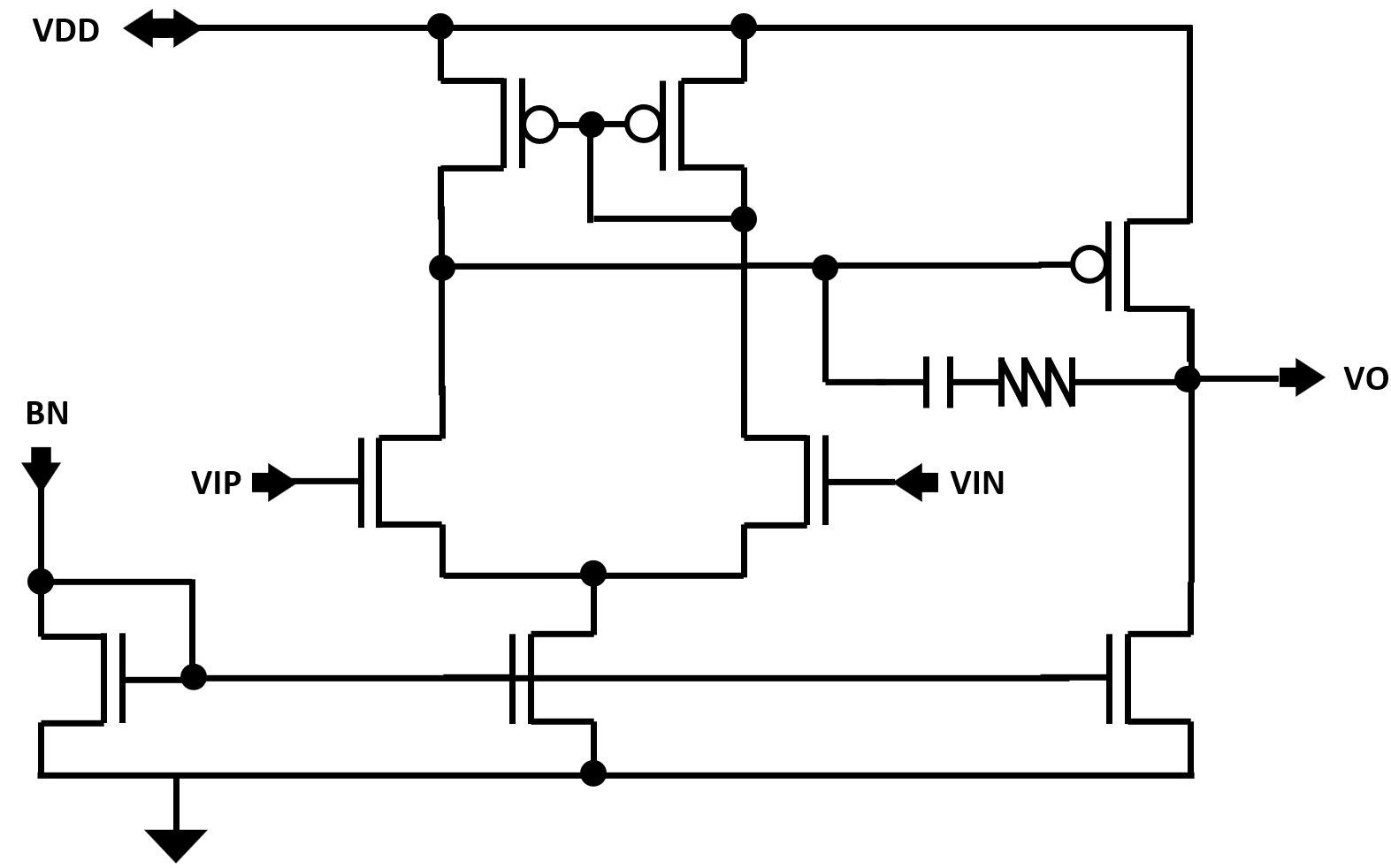}
    \caption{Two-stage operational amplifier schematic}
    \label{fig:two_stage_schematic}
\end{figure}

\begin{figure}[!t]
    \centering
    \includegraphics[trim=0 10 0 50,width=1.0\linewidth]{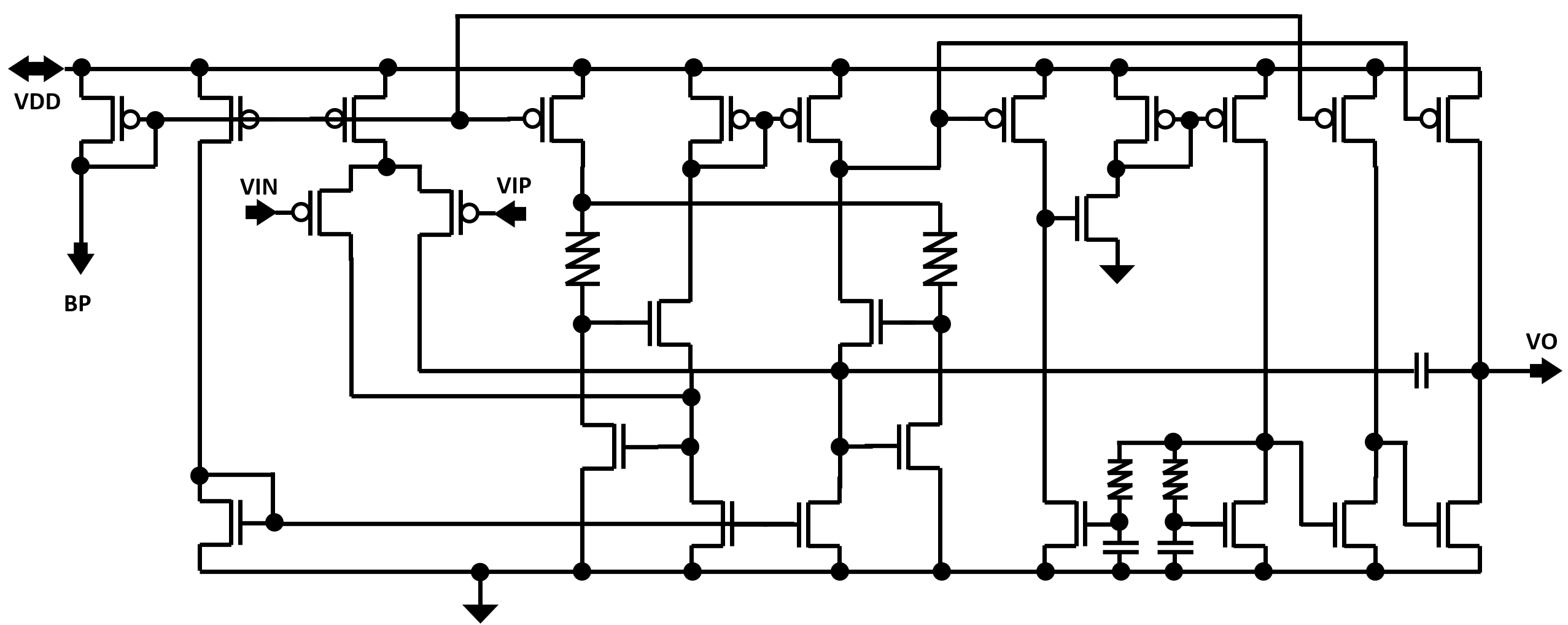}
    \caption{Multi-stage operational amplifier schematic}
    \label{fig:multi_stage_schematic}
    \vspace{-0.3cm}
\end{figure}

We evaluated Lighthouse RL on two analog circuits.
The first is a two-stage operational amplifier~(Fig.~\ref{fig:two_stage_schematic}) consisting of eight transistors, one compensation capacitor, and one resistor. Due to matching requirements, certain transistors share parameters, resulting in 11 independent variables that need to be optimized.
The second circuit is a more complex multistage amplifier from~\cite{qu2016design_big_opamp}~(Fig.~\ref{fig:multi_stage_schematic})
that is provided as benchmark circuit in~\cite{analog_gym} with 29 independent parameters.
This represents one of the largest parameter spaces addressed in the RL-based circuit sizing literature~(Table~\ref{tab:rl_circuit_params}).
The decision parameters and the ranges of training objectives are detailed in Table~\ref{tab:design_spec_ranges}.

For our comparative analysis, we incorporate several baseline methods: (1) Bayesian Optimization~\cite{shahriari2015taking_bo_simple}, a training-free black-box optimization approach; (2) RL Backbone, our foundational reinforcement learning system, which starts episodes from a fixed point; (3) RoSeOpt~\cite{roseopt_rl1}, a state-of-the-art circuit optimization method that combines BO for initialization with RL for optimization, which has previously demonstrated superior performance compared to various optimization methods including genetic algorithms~\cite{liu2009analog_es3}, Bayesian optimization~\cite{lyu2018bo1}, and other RL-based circuit optimization methods~\cite{settaluri2021automated_rl4, shi2022robustanalog_rl2, wang2020gcn_rl5}; and (4) RoSeOpt with fine-grained action space, our modification that uses smaller discretization steps~(0.25× the original for two-stage OpAmp and 0.5× for multistage amplifier) for more precise parameter adjustments, as we observed the original discretization in the paper caused abrupt objective changes for our target circuits. We evaluate each method with open-source Skywater SKY130 PDK~\cite{skywater130} with Ngspice~\cite{ngspice} for simulation and with a typical process corner~(TT).

For fair comparison, all learning-based methods were allocated 6,000 SPICE~\cite{nagel1973spice} simulations for the two-stage OpAmp and 30,000 simulations for the multistage amplifier during training. For the first two experiments, all methods were evaluated on 50 different target objectives sampled from the ranges in Table~\ref{tab:design_spec_ranges}, with a maximum of 30 inference steps allowed per target. Since BO does not require training, we allocate it a total budget equivalent to the combined training and inference budgets of other methods~(e.g., 7,500 simulations, that are 150 simulations per target) for the two-stage OpAmp: 6,000 training + 50 targets × 30 steps maximum, similarly 630 simulations/target for multistage amplifier). Each method was evaluated in five different random seeds for reproducibility.

\begin{figure}[!t]
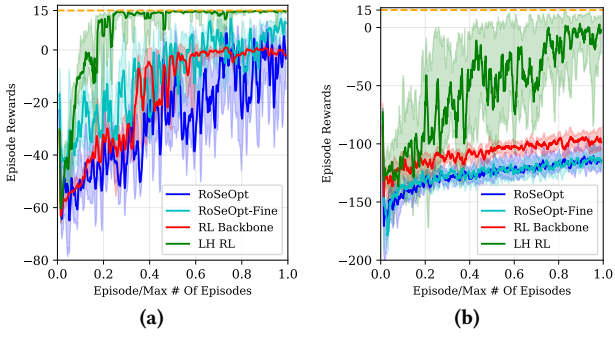

    \centering
    \begin{subfigure}[t]{0.23\textwidth}
        \centering
        \includesvg[width=\linewidth]{figures/episode_rewards_median_iqr.svg}
        \vspace{-0.6cm}
        \caption{}
        \label{fig:reward_comparison_three_methods}
    \end{subfigure}
    \begin{subfigure}[t]{0.23\textwidth}
        \centering
        \includesvg[width=\linewidth]{figures/episode_rewards_median_iqr_2.svg}
        \vspace{-0.6cm}
        \caption{}
        \label{fig:reward_comparison_three_methods_2}
    \end{subfigure}
    \vspace{-0.2cm}
    \caption{Reward performance comparison across methods for (a) two-stage OpAmp (6,000 SPICE simulations) and (b) multistage amplifier (30,000 SPICE simulations): RL backbone (red), RoSeOpt (dark blue), RoSeOpt fine-grained (light blue), and Lighthouse RL (green). The X-axis shows normalized training progress for fair comparison, as methods complete different numbers of episodes within the same simulation budget. Lines show median rewards (moving average, window=10) with IQR shading across five random seeds.}
    \label{fig:combined_circuit}
    \squeezeSpaceFigures
\end{figure}

All methods' training progressions for the first two experiments are shown in Fig.~\ref{fig:reward_comparison_three_methods} and Fig.~\ref{fig:reward_comparison_three_methods_2}, where Lighthouse RL shows faster convergence and higher rewards for both circuits. 

\subsection{Experiment 1: Generalization Within Training Distribution}

Our first experiment tests whether each method can meet diverse target objectives within the training ranges~(Table~\ref{tab:design_spec_ranges}), i.e., finding parameters that satisfy varying gain, bandwidth, phase margin, and gain margin requirements without retraining. As shown in Table~\ref{tab:comprehensive_comparison}, Lighthouse RL outperforms all other methods across every evaluation metric: 

For the two-stage OpAmp, Lighthouse RL achieves a 1.55× improvement in sample efficiency compared to RoSeOpt~(1.00×), requiring fewer circuit simulations for convergence. This advantage grows to 1.72× for the multistage amplifier, compared to RoSeOpt's 1.02×, demonstrating that Lighthouse RL's efficiency scales favorably with problem complexity. As BO does not have a training phase, this metric is not applicable.

For the two-stage OpAmp, Lighthouse RL achieves 100.0\,$\pm$\,0.0\% success rate across all targets and seeds, matching the RL backbone but significantly outperforming both the RoSeOpt variants~(58.2\,$\pm$\,15.4\% and 87.2\,$\pm$\,14.7\% respectively) and BO~(37.6\,$\pm$\,4.8\%). The difference becomes even more pronounced for the multistage amplifier, where Lighthouse RL maintains an 87.2\,$\pm$\,18.0\% success rate versus the BO, which has 26.4\,$\pm$\,6.2.

For the two-stage OpAmp, Lighthouse RL requires only 3.8\,$\pm$\,1.4 steps on average during inference, representing a 4.1$\times$ reduction compared to the RL backbone and a 2.9-4.7$\times$ reduction compared to the RoSeOpt variants. For the multistage amplifier, Lighthouse RL requires only 7.9\,$\pm$\,4.85 steps, compared to nearly 30 steps for all learning-based methods, resulting in a 3.5-3.8$\times$ improvement. 
This reduction is relevant for circuit sizing, where simulations are computationally expensive.

Lighthouse RL outperforms benchmarks by strategically using lighthouse states as starting points, enabling efficient navigation to target objectives.  This advantage becomes more pronounced as parameter dimensions increase from 11 to 29, where Lighthouse RL maintains high success rates while other methods struggle. Although one RoSeOpt seed performed exceptionally well, its high variance~(13.2 $\pm$ 26. 4\%) reveals a dependence on the initial parameters discovered by BO. 
Lighthouse RL shows better robustness by continually exploiting multiple high-performing states during training.

\subsection{Experiment 2: Extrapolation Beyond Training Distribution}

Although there is good performance within the training distribution, the true value of Lighthouse RL lies in its ability to extrapolate to objectives beyond those seen during training. This extrapolation capability is especially significant, as otherwise we could simply provide designers with the best parameter sets found during training. The extrapolation task reveals even more pronounced differences between the methods.

For the two-stage OpAmp, Lighthouse RL achieves 75.2\,±\,26.8\% success rate when extrapolating to more demanding objectives,
significantly outperforming BO~(8.0\,±\,4.38\%), both RoSeOpt variants~(5.6\,±\,4.0\% and 10.0\,±\,14.4\%), and the RL backbone~(50.4\,±\,41.6\%).
Lighthouse RL requires only 15.6\,±\,6.5 inference steps on average,
that is 9.3× and 1.6-1.8× improvement over BO and other RL methods, respectively.

For the multistage amplifier, Lighthouse RL outperforms benchmarks with a 29.2\,±\,37.0\% success rate versus BO's 10.0\,±\,1.8\%.
This highlights how Lighthouse RL's strategic reset
is increasingly valuable as problem complexity grows.

\subsection{Experiment 3: Objective Maximization}

\begin{table}[!t]
\centering
\caption{Objective maximization results for multistage amplifier. Best results are shown in \textbf{bold}.}
\label{tab:spec_max_exp}
\setlength{\tabcolsep}{5pt}
\renewcommand{\arraystretch}{1.0}
\begin{tabular}{@{}lcccc@{}}
\toprule
\textbf{Method} & \textbf{Gain} & \textbf{BW} & \textbf{PM} & \textbf{GM} \\
 & \textbf{(dB)} & \textbf{(MHz)} & \textbf{(°)} & \textbf{(dB)} \\
\midrule
Human Design & 91 & 1.2 & 86.3 & 57.8 \\
RL Backbone & 52.5 & 13.1 & 59.0 & 15.9 \\
LH RL & \textbf{118.1} & \textbf{24.5} & \textbf{89.7} & \textbf{59.6} \\
\bottomrule
\end{tabular}
\vspace{-0.3cm}
\end{table}

The last experiment focuses on maximizing individual circuit objectives rather than meeting specific targets in a range, a common design scenario where engineers seek to push performance limits. We conduct this experiment on the multistage amplifier. 

We modify our approach by providing unattainable target objectives~(Table~\ref{tab:design_spec_ranges}) during training. For this experiment, we only implement Phase 1 of Algorithm~\ref{alg:sac_reset}, continuously updating the lighthouse states rather than fixing them. This approach ensures that we always maintain and improve the best-performing configurations discovered during training.
Each method is trained for 30,000~time steps for 3 different seeds and the solution with highest reward is reported.

We compare Lighthouse RL against a human designer and the RL backbone. As shown in Table~\ref{tab:spec_max_exp}, Lighthouse RL outperforms all baselines in all objectives. It achieves a gain of 118.1\,dB, a bandwidth of 24.5\,MHz, a phase margin of 89.7°, and a gain margin of 59.6\,dB.
Note that no constraints on current consumption were imposed, enabling Lighthouse RL to achieve the reported results; additional constraints may be required in some practical settings.
 
\section{Conclusion}

In this paper, we introduced Lighthouse RL, a novel reinforcement learning approach for efficient black-box optimization that leverages strategic reset points to guide exploration. By maintaining a collection of high-performing parameter configurations discovered during training, our method significantly improves sample efficiency and generalization capabilities compared to conventional approaches. We demonstrated the effectiveness of our approach on a benchmark optimization problem and practical circuit sizing tasks, showing substantial improvements in success rates, inference efficiency, and objective maximization. The proposed reset strategy can be integrated with any RL-based optimization approach, making it valuable for addressing complex black-box optimization problems. Limitations remain in lighthouse state selection, as their diversity currently depends on training dynamics rather than explicit diversity enforcement. Future work could incorporate explicit diversity metrics and explore parallelization to accelerate the discovery of high-performance configurations.

\bibliographystyle{ACM-Reference-Format}
\bibliography{references} 


\begin{thebibliography}{34}


\ifx \showCODEN    \undefined \def \showCODEN     #1{\unskip}     \fi
\ifx \showDOI      \undefined \def \showDOI       #1{#1}\fi
\ifx \showISBNx    \undefined \def \showISBNx     #1{\unskip}     \fi
\ifx \showISBNxiii \undefined \def \showISBNxiii  #1{\unskip}     \fi
\ifx \showISSN     \undefined \def \showISSN      #1{\unskip}     \fi
\ifx \showLCCN     \undefined \def \showLCCN      #1{\unskip}     \fi
\ifx \shownote     \undefined \def \shownote      #1{#1}          \fi
\ifx \showarticletitle \undefined \def \showarticletitle #1{#1}   \fi
\ifx \showURL      \undefined \def \showURL       {\relax}        \fi
\providecommand\bibfield[2]{#2}
\providecommand\bibinfo[2]{#2}
\providecommand\natexlab[1]{#1}
\providecommand\showeprint[2][]{arXiv:#2}

\bibitem[Andrychowicz et~al\mbox{.}(2017)]%
        {andrychowicz2017hindsight}
\bibfield{author}{\bibinfo{person}{Marcin Andrychowicz}, \bibinfo{person}{Filip Wolski}, \bibinfo{person}{Alex Ray}, \bibinfo{person}{Jonas Schneider}, \bibinfo{person}{Rachel Fong}, \bibinfo{person}{Peter Welinder}, \bibinfo{person}{Bob McGrew}, \bibinfo{person}{Josh Tobin}, \bibinfo{person}{OpenAI Pieter~Abbeel}, {and} \bibinfo{person}{Wojciech Zaremba}.} \bibinfo{year}{2017}\natexlab{}.
\newblock \showarticletitle{Hindsight experience replay}.
\newblock \bibinfo{journal}{\emph{Advances in neural information processing systems}}  \bibinfo{volume}{30} (\bibinfo{year}{2017}).
\newblock


\bibitem[Ba et~al\mbox{.}(2016)]%
        {ba2016layernorm}
\bibfield{author}{\bibinfo{person}{Jimmy~Lei Ba}, \bibinfo{person}{Jamie~Ryan Kiros}, {and} \bibinfo{person}{Geoffrey~E Hinton}.} \bibinfo{year}{2016}\natexlab{}.
\newblock \showarticletitle{Layer normalization}.
\newblock \bibinfo{journal}{\emph{arXiv preprint arXiv:1607.06450}} (\bibinfo{year}{2016}).
\newblock


\bibitem[Beyer and Schwefel(2002)]%
        {beyer2002evolution_es1}
\bibfield{author}{\bibinfo{person}{Hans-Georg Beyer} {and} \bibinfo{person}{Hans-Paul Schwefel}.} \bibinfo{year}{2002}\natexlab{}.
\newblock \showarticletitle{Evolution strategies--a comprehensive introduction}.
\newblock \bibinfo{journal}{\emph{Natural computing}} \bibinfo{volume}{1}, \bibinfo{number}{1} (\bibinfo{year}{2002}), \bibinfo{pages}{3--52}.
\newblock


\bibitem[Budak et~al\mbox{.}(2021)]%
        {budak2021dnn_rl3}
\bibfield{author}{\bibinfo{person}{Ahmet~F Budak}, \bibinfo{person}{Prateek Bhansali}, \bibinfo{person}{Bo Liu}, \bibinfo{person}{Nan Sun}, \bibinfo{person}{David~Z Pan}, {and} \bibinfo{person}{Chandramouli~V Kashyap}.} \bibinfo{year}{2021}\natexlab{}.
\newblock \showarticletitle{Dnn-opt: An rl inspired optimization for analog circuit sizing using deep neural networks}. In \bibinfo{booktitle}{\emph{2021 58th ACM/IEEE Design Automation Conference (DAC)}}. IEEE, \bibinfo{pages}{1219--1224}.
\newblock


\bibitem[Cao et~al\mbox{.}(2024)]%
        {roseopt_rl1}
\bibfield{author}{\bibinfo{person}{Weidong Cao}, \bibinfo{person}{Jian Gao}, \bibinfo{person}{Tianrui Ma}, \bibinfo{person}{Rui Ma}, \bibinfo{person}{Mouhacine Benosman}, {and} \bibinfo{person}{Xuan Zhang}.} \bibinfo{year}{2024}\natexlab{}.
\newblock \showarticletitle{Rose-opt: Robust and efficient analog circuit parameter optimization with knowledge-infused reinforcement learning}.
\newblock \bibinfo{journal}{\emph{IEEE Transactions on Computer-Aided Design of Integrated Circuits and Systems}} (\bibinfo{year}{2024}).
\newblock


\bibitem[Fujimoto et~al\mbox{.}(2018)]%
        {fujimoto2018addressingtd3}
\bibfield{author}{\bibinfo{person}{Scott Fujimoto}, \bibinfo{person}{Herke Hoof}, {and} \bibinfo{person}{David Meger}.} \bibinfo{year}{2018}\natexlab{}.
\newblock \showarticletitle{Addressing function approximation error in actor-critic methods}. In \bibinfo{booktitle}{\emph{International conference on machine learning}}. PMLR, \bibinfo{pages}{1587--1596}.
\newblock


\bibitem[Google and Foundry(2020)]%
        {skywater130}
\bibfield{author}{\bibinfo{person}{Google} {and} \bibinfo{person}{SkyWater~Technology Foundry}.} \bibinfo{year}{2020}\natexlab{}.
\newblock \bibinfo{title}{SkyWater 130nm PDK}.
\newblock \bibinfo{howpublished}{\url{https://github.com/google/skywater-pdk}}.
\newblock
\newblock
\shownote{[Online]}.


\bibitem[Haarnoja et~al\mbox{.}(2018a)]%
        {haarnoja2018soft}
\bibfield{author}{\bibinfo{person}{Tuomas Haarnoja}, \bibinfo{person}{Aurick Zhou}, \bibinfo{person}{Pieter Abbeel}, {and} \bibinfo{person}{Sergey Levine}.} \bibinfo{year}{2018}\natexlab{a}.
\newblock \showarticletitle{Soft actor-critic: Off-policy maximum entropy deep reinforcement learning with a stochastic actor}. In \bibinfo{booktitle}{\emph{International conference on machine learning}}. Pmlr, \bibinfo{pages}{1861--1870}.
\newblock


\bibitem[Haarnoja et~al\mbox{.}(2018b)]%
        {haarnoja2018sactheoryandapplications}
\bibfield{author}{\bibinfo{person}{Tuomas Haarnoja}, \bibinfo{person}{Aurick Zhou}, \bibinfo{person}{Kristian Hartikainen}, \bibinfo{person}{George Tucker}, \bibinfo{person}{Sehoon Ha}, \bibinfo{person}{Jie Tan}, \bibinfo{person}{Vikash Kumar}, \bibinfo{person}{Henry Zhu}, \bibinfo{person}{Abhishek Gupta}, \bibinfo{person}{Pieter Abbeel}, {et~al\mbox{.}}} \bibinfo{year}{2018}\natexlab{b}.
\newblock \showarticletitle{Soft actor-critic algorithms and applications}.
\newblock \bibinfo{journal}{\emph{arXiv preprint arXiv:1812.05905}} (\bibinfo{year}{2018}).
\newblock


\bibitem[Hakhamaneshi et~al\mbox{.}(2019)]%
        {hakhamaneshi2019bagnet_es2}
\bibfield{author}{\bibinfo{person}{Kourosh Hakhamaneshi}, \bibinfo{person}{Nick Werblun}, \bibinfo{person}{Pieter Abbeel}, {and} \bibinfo{person}{Vladimir Stojanovi{\'c}}.} \bibinfo{year}{2019}\natexlab{}.
\newblock \showarticletitle{BagNet: Berkeley analog generator with layout optimizer boosted with deep neural networks}. In \bibinfo{booktitle}{\emph{2019 IEEE/ACM International Conference on Computer-Aided Design (ICCAD)}}. IEEE, \bibinfo{pages}{1--8}.
\newblock


\bibitem[Kiran et~al\mbox{.}(2021)]%
        {kiran2021deep_autonomous_driving}
\bibfield{author}{\bibinfo{person}{B~Ravi Kiran}, \bibinfo{person}{Ibrahim Sobh}, \bibinfo{person}{Victor Talpaert}, \bibinfo{person}{Patrick Mannion}, \bibinfo{person}{Ahmad~A Al~Sallab}, \bibinfo{person}{Senthil Yogamani}, {and} \bibinfo{person}{Patrick P{\'e}rez}.} \bibinfo{year}{2021}\natexlab{}.
\newblock \showarticletitle{Deep reinforcement learning for autonomous driving: A survey}.
\newblock \bibinfo{journal}{\emph{IEEE transactions on intelligent transportation systems}} \bibinfo{volume}{23}, \bibinfo{number}{6} (\bibinfo{year}{2021}), \bibinfo{pages}{4909--4926}.
\newblock


\bibitem[Li et~al\mbox{.}(2025)]%
        {analog_gym}
\bibfield{author}{\bibinfo{person}{Jintao Li}, \bibinfo{person}{Haochang Zhi}, \bibinfo{person}{Ruiyu Lyu}, \bibinfo{person}{Wangzhen Li}, \bibinfo{person}{Zhaori Bi}, \bibinfo{person}{Keren Zhu}, \bibinfo{person}{Yanhan Zeng}, \bibinfo{person}{Weiwei Shan}, \bibinfo{person}{Changhao Yan}, \bibinfo{person}{Fan Yang}, \bibinfo{person}{Yun Li}, {and} \bibinfo{person}{Xuan Zeng}.} \bibinfo{year}{2025}\natexlab{}.
\newblock \showarticletitle{AnalogGym: An Open and Practical Testing Suite for Analog Circuit Synthesis}. In \bibinfo{booktitle}{\emph{Proceedings of the 43rd IEEE/ACM International Conference on Computer-Aided Design}} (Newark Liberty International Airport Marriott, New York, NY, USA) \emph{(\bibinfo{series}{ICCAD '24})}. \bibinfo{publisher}{Association for Computing Machinery}, \bibinfo{address}{New York, NY, USA}, Article \bibinfo{articleno}{59}, \bibinfo{numpages}{9}~pages.
\newblock
\showISBNx{9798400710773}
\urldef\tempurl%
\url{https://doi.org/10.1145/3676536.3697117}
\showDOI{\tempurl}


\bibitem[Lillicrap et~al\mbox{.}(2015)]%
        {lillicrap2015continuousddpg}
\bibfield{author}{\bibinfo{person}{Timothy~P Lillicrap}, \bibinfo{person}{Jonathan~J Hunt}, \bibinfo{person}{Alexander Pritzel}, \bibinfo{person}{Nicolas Heess}, \bibinfo{person}{Tom Erez}, \bibinfo{person}{Yuval Tassa}, \bibinfo{person}{David Silver}, {and} \bibinfo{person}{Daan Wierstra}.} \bibinfo{year}{2015}\natexlab{}.
\newblock \showarticletitle{Continuous control with deep reinforcement learning}.
\newblock \bibinfo{journal}{\emph{arXiv preprint arXiv:1509.02971}} (\bibinfo{year}{2015}).
\newblock


\bibitem[Liu et~al\mbox{.}(2009)]%
        {liu2009analog_es3}
\bibfield{author}{\bibinfo{person}{Bo Liu}, \bibinfo{person}{Yan Wang}, \bibinfo{person}{Zhiping Yu}, \bibinfo{person}{Leibo Liu}, \bibinfo{person}{Miao Li}, \bibinfo{person}{Zheng Wang}, \bibinfo{person}{Jing Lu}, {and} \bibinfo{person}{Francisco~V Fern{\'a}ndez}.} \bibinfo{year}{2009}\natexlab{}.
\newblock \showarticletitle{Analog circuit optimization system based on hybrid evolutionary algorithms}.
\newblock \bibinfo{journal}{\emph{Integration}} \bibinfo{volume}{42}, \bibinfo{number}{2} (\bibinfo{year}{2009}), \bibinfo{pages}{137--148}.
\newblock


\bibitem[Liu et~al\mbox{.}(2021)]%
        {liu2021parasiticbo4}
\bibfield{author}{\bibinfo{person}{Mingjie Liu}, \bibinfo{person}{Walker~J Turner}, \bibinfo{person}{George~F Kokai}, \bibinfo{person}{Brucek Khailany}, \bibinfo{person}{David~Z Pan}, {and} \bibinfo{person}{Haoxing Ren}.} \bibinfo{year}{2021}\natexlab{}.
\newblock \showarticletitle{Parasitic-aware analog circuit sizing with graph neural networks and Bayesian optimization}. In \bibinfo{booktitle}{\emph{2021 Design, automation \& test in Europe conference \& exhibition (DATE)}}. IEEE, \bibinfo{pages}{1372--1377}.
\newblock


\bibitem[Lyu et~al\mbox{.}(2017)]%
        {lyu2017bo2}
\bibfield{author}{\bibinfo{person}{Wenlong Lyu}, \bibinfo{person}{Pan Xue}, \bibinfo{person}{Fan Yang}, \bibinfo{person}{Changhao Yan}, \bibinfo{person}{Zhiliang Hong}, \bibinfo{person}{Xuan Zeng}, {and} \bibinfo{person}{Dian Zhou}.} \bibinfo{year}{2017}\natexlab{}.
\newblock \showarticletitle{An efficient Bayesian optimization approach for automated optimization of analog circuits}.
\newblock \bibinfo{journal}{\emph{IEEE Transactions on Circuits and Systems I: Regular Papers}} \bibinfo{volume}{65}, \bibinfo{number}{6} (\bibinfo{year}{2017}), \bibinfo{pages}{1954--1967}.
\newblock


\bibitem[Lyu et~al\mbox{.}(2018)]%
        {lyu2018bo1}
\bibfield{author}{\bibinfo{person}{Wenlong Lyu}, \bibinfo{person}{Fan Yang}, \bibinfo{person}{Changhao Yan}, \bibinfo{person}{Dian Zhou}, {and} \bibinfo{person}{Xuan Zeng}.} \bibinfo{year}{2018}\natexlab{}.
\newblock \showarticletitle{Batch Bayesian optimization via multi-objective acquisition ensemble for automated analog circuit design}. In \bibinfo{booktitle}{\emph{International conference on machine learning}}. PMLR, \bibinfo{pages}{3306--3314}.
\newblock


\bibitem[Mnih et~al\mbox{.}(2013)]%
        {mnih2013playingatarideepreinforcement_atari}
\bibfield{author}{\bibinfo{person}{Volodymyr Mnih}, \bibinfo{person}{Koray Kavukcuoglu}, \bibinfo{person}{David Silver}, \bibinfo{person}{Alex Graves}, \bibinfo{person}{Ioannis Antonoglou}, \bibinfo{person}{Daan Wierstra}, {and} \bibinfo{person}{Martin Riedmiller}.} \bibinfo{year}{2013}\natexlab{}.
\newblock \bibinfo{title}{Playing Atari with Deep Reinforcement Learning}.
\newblock
\newblock
\showeprint[arxiv]{1312.5602}~[cs.LG]
\urldef\tempurl%
\url{https://arxiv.org/abs/1312.5602}
\showURL{%
\tempurl}


\bibitem[Nagel and Pederson(1973)]%
        {nagel1973spice}
\bibfield{author}{\bibinfo{person}{Laurence Nagel} {and} \bibinfo{person}{Donald~O Pederson}.} \bibinfo{year}{1973}\natexlab{}.
\newblock \showarticletitle{SPICE (simulation program with integrated circuit emphasis)}.
\newblock  (\bibinfo{year}{1973}).
\newblock


\bibitem[{Ngspice Contributors}(2025)]%
        {ngspice}
\bibfield{author}{\bibinfo{person}{{Ngspice Contributors}}.} \bibinfo{year}{2025}\natexlab{}.
\newblock \bibinfo{title}{Ngspice, An Open-Source SPICE Simulator}.
\newblock \bibinfo{howpublished}{\url{http://ngspice.sourceforge.net/}}.
\newblock
\newblock
\shownote{[Online]}.


\bibitem[Qu et~al\mbox{.}(2016)]%
        {qu2016design_big_opamp}
\bibfield{author}{\bibinfo{person}{Wanyuan Qu}, \bibinfo{person}{Shashank Singh}, \bibinfo{person}{Yongjin Lee}, \bibinfo{person}{Young-Suk Son}, {and} \bibinfo{person}{Gyu-Hyeong Cho}.} \bibinfo{year}{2016}\natexlab{}.
\newblock \showarticletitle{Design-oriented analysis for Miller compensation and its application to multistage amplifier design}.
\newblock \bibinfo{journal}{\emph{IEEE Journal of Solid-State Circuits}} \bibinfo{volume}{52}, \bibinfo{number}{2} (\bibinfo{year}{2016}), \bibinfo{pages}{517--527}.
\newblock


\bibitem[Schulman et~al\mbox{.}(2015)]%
        {schulman2015trust}
\bibfield{author}{\bibinfo{person}{John Schulman}, \bibinfo{person}{Sergey Levine}, \bibinfo{person}{Pieter Abbeel}, \bibinfo{person}{Michael Jordan}, {and} \bibinfo{person}{Philipp Moritz}.} \bibinfo{year}{2015}\natexlab{}.
\newblock \showarticletitle{Trust region policy optimization}. In \bibinfo{booktitle}{\emph{International conference on machine learning}}. PMLR, \bibinfo{pages}{1889--1897}.
\newblock


\bibitem[Schulman et~al\mbox{.}(2017)]%
        {schulman2017proximal}
\bibfield{author}{\bibinfo{person}{John Schulman}, \bibinfo{person}{Filip Wolski}, \bibinfo{person}{Prafulla Dhariwal}, \bibinfo{person}{Alec Radford}, {and} \bibinfo{person}{Oleg Klimov}.} \bibinfo{year}{2017}\natexlab{}.
\newblock \showarticletitle{Proximal policy optimization algorithms}.
\newblock \bibinfo{journal}{\emph{arXiv preprint arXiv:1707.06347}} (\bibinfo{year}{2017}).
\newblock


\bibitem[Settaluri et~al\mbox{.}(2021)]%
        {settaluri2021automated_rl4}
\bibfield{author}{\bibinfo{person}{Keertana Settaluri}, \bibinfo{person}{Zhaokai Liu}, \bibinfo{person}{Rishubh Khurana}, \bibinfo{person}{Arash Mirhaj}, \bibinfo{person}{Rajeev Jain}, {and} \bibinfo{person}{Borivoje Nikolic}.} \bibinfo{year}{2021}\natexlab{}.
\newblock \showarticletitle{Automated design of analog circuits using reinforcement learning}.
\newblock \bibinfo{journal}{\emph{IEEE Transactions on Computer-Aided Design of Integrated Circuits and Systems}} \bibinfo{volume}{41}, \bibinfo{number}{9} (\bibinfo{year}{2021}), \bibinfo{pages}{2794--2807}.
\newblock


\bibitem[Shahriari et~al\mbox{.}(2015)]%
        {shahriari2015taking_bo_simple}
\bibfield{author}{\bibinfo{person}{Bobak Shahriari}, \bibinfo{person}{Kevin Swersky}, \bibinfo{person}{Ziyu Wang}, \bibinfo{person}{Ryan~P Adams}, {and} \bibinfo{person}{Nando De~Freitas}.} \bibinfo{year}{2015}\natexlab{}.
\newblock \showarticletitle{Taking the human out of the loop: A review of Bayesian optimization}.
\newblock \bibinfo{journal}{\emph{Proc. IEEE}} \bibinfo{volume}{104}, \bibinfo{number}{1} (\bibinfo{year}{2015}), \bibinfo{pages}{148--175}.
\newblock


\bibitem[Shi et~al\mbox{.}(2022)]%
        {shi2022robustanalog_rl2}
\bibfield{author}{\bibinfo{person}{Wei Shi}, \bibinfo{person}{Hanrui Wang}, \bibinfo{person}{Jiaqi Gu}, \bibinfo{person}{Mingjie Liu}, \bibinfo{person}{David~Z Pan}, \bibinfo{person}{Song Han}, {and} \bibinfo{person}{Nan Sun}.} \bibinfo{year}{2022}\natexlab{}.
\newblock \showarticletitle{RobustAnalog: Fast variation-aware analog circuit design via multi-task RL}. In \bibinfo{booktitle}{\emph{Proceedings of the 2022 ACM/IEEE Workshop on Machine Learning for CAD}}. \bibinfo{pages}{35--41}.
\newblock


\bibitem[Sutton et~al\mbox{.}(1998)]%
        {sutton1998reinforcement}
\bibfield{author}{\bibinfo{person}{Richard~S Sutton}, \bibinfo{person}{Andrew~G Barto}, {et~al\mbox{.}}} \bibinfo{year}{1998}\natexlab{}.
\newblock \bibinfo{booktitle}{\emph{Reinforcement learning: An introduction}}. Vol.~\bibinfo{volume}{1}.
\newblock \bibinfo{publisher}{MIT press Cambridge}.
\newblock


\bibitem[Sutton et~al\mbox{.}(1999)]%
        {sutton1999policyREINFORCE}
\bibfield{author}{\bibinfo{person}{Richard~S Sutton}, \bibinfo{person}{David McAllester}, \bibinfo{person}{Satinder Singh}, {and} \bibinfo{person}{Yishay Mansour}.} \bibinfo{year}{1999}\natexlab{}.
\newblock \showarticletitle{Policy gradient methods for reinforcement learning with function approximation}.
\newblock \bibinfo{journal}{\emph{Advances in neural information processing systems}}  \bibinfo{volume}{12} (\bibinfo{year}{1999}).
\newblock


\bibitem[Tang et~al\mbox{.}(2025)]%
        {tang2025deep_robotics}
\bibfield{author}{\bibinfo{person}{Chen Tang}, \bibinfo{person}{Ben Abbatematteo}, \bibinfo{person}{Jiaheng Hu}, \bibinfo{person}{Rohan Chandra}, \bibinfo{person}{Roberto Mart{\'\i}n-Mart{\'\i}n}, {and} \bibinfo{person}{Peter Stone}.} \bibinfo{year}{2025}\natexlab{}.
\newblock \showarticletitle{Deep reinforcement learning for robotics: A survey of real-world successes}. In \bibinfo{booktitle}{\emph{Proceedings of the AAAI Conference on Artificial Intelligence}}, Vol.~\bibinfo{volume}{39}. \bibinfo{pages}{28694--28698}.
\newblock


\bibitem[Touloupas et~al\mbox{.}(2021)]%
        {touloupas2021bo3}
\bibfield{author}{\bibinfo{person}{Konstantinos Touloupas}, \bibinfo{person}{Nikos Chouridis}, {and} \bibinfo{person}{Paul~P Sotiriadis}.} \bibinfo{year}{2021}\natexlab{}.
\newblock \showarticletitle{Local Bayesian optimization for analog circuit sizing}. In \bibinfo{booktitle}{\emph{2021 58th ACM/IEEE design automation conference (DAC)}}. IEEE, \bibinfo{pages}{1237--1242}.
\newblock


\bibitem[Touloupas and Sotiriadis(2021)]%
        {touloupas2021locomobo_bo5}
\bibfield{author}{\bibinfo{person}{Konstantinos Touloupas} {and} \bibinfo{person}{Paul~P Sotiriadis}.} \bibinfo{year}{2021}\natexlab{}.
\newblock \showarticletitle{LoCoMOBO: A local constrained multiobjective Bayesian optimization for analog circuit sizing}.
\newblock \bibinfo{journal}{\emph{IEEE Transactions on Computer-Aided Design of Integrated Circuits and Systems}} \bibinfo{volume}{41}, \bibinfo{number}{9} (\bibinfo{year}{2021}), \bibinfo{pages}{2780--2793}.
\newblock


\bibitem[Uhlmann et~al\mbox{.}(2022)]%
        {uhlmann2022deep_rl6}
\bibfield{author}{\bibinfo{person}{Yannick Uhlmann}, \bibinfo{person}{Michael Essich}, \bibinfo{person}{Lennart Bramlage}, \bibinfo{person}{J{\"u}rgen Scheible}, {and} \bibinfo{person}{Crist{\'o}bal Curio}.} \bibinfo{year}{2022}\natexlab{}.
\newblock \showarticletitle{Deep reinforcement learning for analog circuit sizing with an electrical design space and sparse rewards}. In \bibinfo{booktitle}{\emph{Proceedings of the 2022 ACM/IEEE Workshop on Machine Learning for CAD}}. \bibinfo{pages}{21--26}.
\newblock


\bibitem[Veli{\v{c}}kovi{\'c} et~al\mbox{.}(2017)]%
        {velivckovic2017graph_gat}
\bibfield{author}{\bibinfo{person}{Petar Veli{\v{c}}kovi{\'c}}, \bibinfo{person}{Guillem Cucurull}, \bibinfo{person}{Arantxa Casanova}, \bibinfo{person}{Adriana Romero}, \bibinfo{person}{Pietro Lio}, {and} \bibinfo{person}{Yoshua Bengio}.} \bibinfo{year}{2017}\natexlab{}.
\newblock \showarticletitle{Graph attention networks}.
\newblock \bibinfo{journal}{\emph{arXiv preprint arXiv:1710.10903}} (\bibinfo{year}{2017}).
\newblock


\bibitem[Wang et~al\mbox{.}(2020)]%
        {wang2020gcn_rl5}
\bibfield{author}{\bibinfo{person}{Hanrui Wang}, \bibinfo{person}{Kuan Wang}, \bibinfo{person}{Jiacheng Yang}, \bibinfo{person}{Linxiao Shen}, \bibinfo{person}{Nan Sun}, \bibinfo{person}{Hae-Seung Lee}, {and} \bibinfo{person}{Song Han}.} \bibinfo{year}{2020}\natexlab{}.
\newblock \showarticletitle{GCN-RL circuit designer: Transferable transistor sizing with graph neural networks and reinforcement learning}. In \bibinfo{booktitle}{\emph{2020 57th ACM/IEEE Design Automation Conference (DAC)}}. IEEE, \bibinfo{pages}{1--6}.
\newblock


\end{thebibliography}

\end{document}